\documentclass[letterpaper, 10 pt, conference]{ieeeconf}
\IEEEoverridecommandlockouts
\usepackage{times}
\usepackage{helvet}
\usepackage{courier}
\usepackage[hyphens]{url}
\usepackage{graphicx}
\usepackage{amsmath}
\usepackage{amssymb}
\usepackage{upgreek}
\usepackage{color}
\usepackage{url}
\usepackage{cite}
\usepackage{booktabs}
\usepackage{graphicx}
\usepackage{multirow}
\usepackage{diagbox}
\usepackage{bm}
\usepackage{cases}
\usepackage{graphicx}
\usepackage{colortbl}
\definecolor{mygray}{gray}{.9}
\definecolor{mypink}{rgb}{.99,.91,.95}
\definecolor{mygreen}{rgb}{.70,.93,.70}
\definecolor{mycyan}{cmyk}{.3,0,0,0}

\title{\LARGE \bf
3DCFS: Fast and Robust Joint 3D Semantic-Instance Segmentation via Coupled Feature Selection
}

\author{Liang Du$^{\dagger1}$, Jingang Tan$^{\dagger2}$, Xiangyang Xue$^{3}$, Lili Chen$^{2}$, \\Hongkai Wen$^{1,4}$, Jianfeng Feng$^{1}$, Jiamao Li$^{2}$ and Xiaolin Zhang$^{2}$
\thanks{$\dagger$ The first two authors contributed equally to this work.}
\thanks{* This work was supported by the 111 Project (NO.B18015), the National Natural Science Foundation of China (No.91630314), the key project of Shanghai Science \& Technology (No.16JC1420402), Shanghai Municipal Science and Technology Major Project (No.2018SHZDZX01) and ZJLab.}
\thanks{$^{1}$Liang Du, Hongkai Wen, Jianfeng Feng are with the Institute of Science and Technology for Brain-Inspired Intelligence, Fudan University, China, Key Laboratory of Computational Neuroscience and Brain-Inspired Intelligence (Fudan University), Ministry of Education, China. {\tt\small duliang@mail.ustc.edu.cn, jffeng@fudan.edu.cn}}
\thanks{$^{2}$Jingang Tan, Lili Chen, Jiamao Li and Xiaolin Zhang are with the Shanghai Institute of Microsystem and Information Technology, Chinese Academy of Sciences, Shanghai, China.}
\thanks{$^{3}$Xiangyang Xue is with the School of Computer Science, Fudan University, China.}
\thanks{$^{4}$Hongkai Wen is with the Department of Computer Science, University of Warwick, UK.}
}

\begin{document}

\maketitle
\thispagestyle{empty}
\pagestyle{empty}

\begin{abstract}

 We propose a novel fast and robust 3D point clouds segmentation framework via coupled feature selection, named 3DCFS, that jointly performs semantic and instance segmentation. Inspired by the human scene perception process, we design a novel coupled feature selection module, named CFSM, that adaptively selects and fuses the reciprocal semantic and instance features from two tasks in a coupled manner. To further boost the performance of the instance segmentation task in our 3DCFS, we investigate a loss function that helps the model learn to balance the magnitudes of the output embedding dimensions during training, which makes calculating the Euclidean distance more reliable and enhances the generalizability of the model. Extensive experiments demonstrate that our 3DCFS outperforms state-of-the-art methods on benchmark datasets in terms of accuracy, speed and computational cost.

\end{abstract}

\section{Introduction}
\label{sec:intro}
3D scene understanding based on LiDAR, RGB-D and stereo cameras has received increasing attention from both academia and industry because of its critical role in robotic scene perception, robotic manipulation and autonomous driving \cite{james2017transferring ,du2018weakly}. Instance and semantic segmentation are the most widely used tasks in this research field. Building on the great success achieved in recent years \cite{he2017mask, de2017semantic, long2015fully, landrieu2018large} for each single task, joint learning methods for both tasks \cite{wang2019associatively, pham2019jsis3d} have opened up a more effective way to improve performance and promote further developments.

The two tasks have some common ground that can be associatively utilized to boost their performance. For example, points with different classes must be from different instances, and points from the same instance must be of the same class.
The simplest but most naive methods to jointly perform instance and semantic segmentation are progressively using the predicted semantic labels to further cluster instances or utilizing the predicted instance results as prior knowledge for semantic segmentation.
Nevertheless, using the unreliable upstream prediction as prior information may affect the downstream task; consequently, such approaches may be suboptimal.
Another approach is to directly combine the high-level features of both tasks to perform information integration \cite{wang2019associatively}.
Although semantic and instance segmentation share the same goal of detecting specific informative regions, they have different individual learning orientations, and some part of the information they contain may be contradictory.
Instance segmentation focuses on extracting point features from different objects to distinguish them, while the features extracted by semantic segmentation are used to classify points with different categories; as a result, the features of both tasks definitely contain different task-oriented parts. Therefore, feature selection is an essential step in the reciprocal process.

\begin{figure}[t]
 \begin{center}
  \includegraphics[width=7cm]{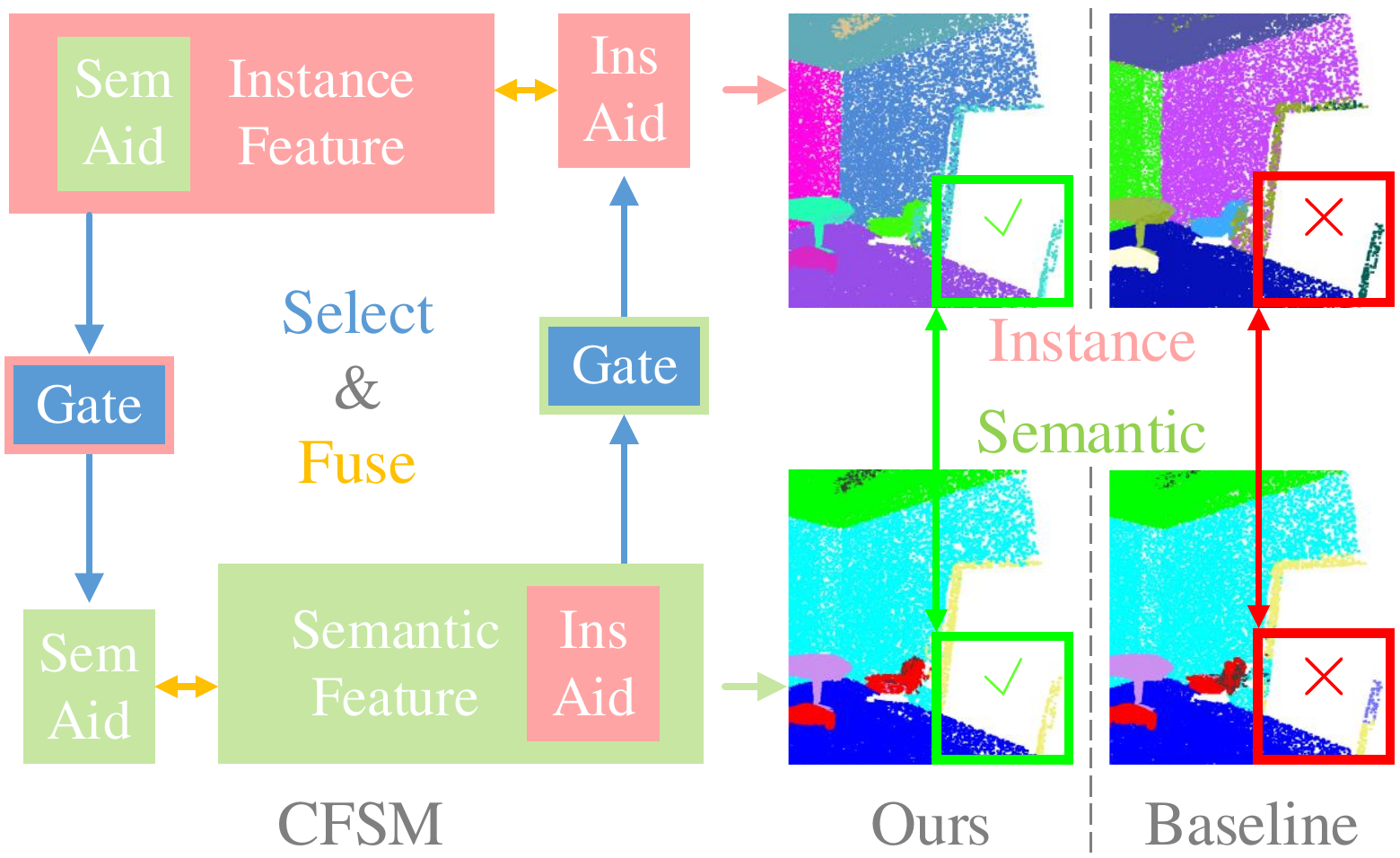}
 \end{center}
 \caption{An illustration of CFSM and a qualitative comparison of our method and the baseline. The baseline is a traditional multitask framework without CFSM, as introduced in Section \ref{sec:method}.
 Our framework via coupled feature selection is able to exploit and integrate the reciprocal information from both tasks based on gate mechanism to boost their performance, as shown in the marked regions.}
 \label{fig1}
\end{figure}

Actually, such selection within a mutually aided process is consistent with human scene perception.
Semantic and instance segmentation are the most important visual tasks in human scene perception.
For humans, semantic perception mainly abstracts the advanced semantic features from the objects in the scene, while instance segmentation pays more attention to exploiting the primary features. These two processes can complement each other. Specifically, the mapping from advanced features to primary features can be beneficial for instance segmentation. For example, if we know the category of an object, we will obtain the blur shape information (primary features), which can help to correct errors in the instance segmentation result if it is difficult to see the whole object because of environmental light. By contrast, if we know the primary features of objects of the same category that are unknown, we can establish links between those that belong to the same category, which can help us to quickly and accurately accomplish the semantic segmentation. Consequently, these two processes are not independent but coupled. However, humans are rarely disturbed by such different task-oriented information because the human brain has the capability to quickly and adaptively select useful information instead of the entire set of information.
This characteristics of human scene perception inspired us to build a multitask-coupled framework to simulate the information-selection process of human scene perception via gate control units for robotics.
The coupled and gate-based training pipeline is shown in Figure \ref{fig1}.
For the encoder, we use the PointNet/PointNet++ utilized by \cite{wang2018sgpn, wang2019associatively}. For the decoder, we investigate a novel coupled feature selection module (CFSM) that contains two coupled instance-to-semantic and semantic-to-instance streams to extract useful information while filtering useless information.

Our 3DCFS uses the Euclidean distance to calculate the similarity of different embeddings among all points for clustering. However, the Euclidean distance is sensitive to the magnitudes of different embedding dimensions, which makes the clustering result depend on only a small number of dimensions and reduces the generalizability of the model. We therefore propose a novel loss function, named $\mathcal{E}_{EMED}$, that helps the model learn to maintain equilibrium among the magnitudes of the instance embedding dimensions.
In summary, the main contributions of our work are as follows:
\begin{itemize}
 \item We propose a fast, yet effective end-to-end point clouds segmentation framework that simultaneously performs semantic and instance segmentation inspired by the human scene perception process.
 \item We introduce a novel coupled feature selection module (CFSM) to exploit the potential reciprocal information in semantic and instance segmentation tasks to seamlessly fuse the heterogeneous features, allowing these two tasks to benefit from each other.
 \item We design a novel loss for instance segmentation in 3DCFS, which helps the model learn to balance the magnitudes of the embedding dimensions to maintain the stability of the Euclidean distance calculation during training.
 \item We achieve state-of-the-art performance for 3D semantic and instance segmentation on benchmark datasets in terms of accuracy, speed and computational cost.
\end{itemize}

\section{Related Work}
\label{sec:related}

\noindent \textbf{2D Semantic and Instance Segmentation.}
The great advances in semantic and instance segmentation have largely been driven by the success of fully convolutional neural networks (FCNs) \cite{long2015fully}.
Numerous approaches \cite{lin2017refinenet, chen2017rethinking, lin2018exploring, li2019dfanet, liu2019structured, he2019adaptive, du2019ssf} based on FCNs have dominated semantic segmentation tasks.
\cite{dai2016instance, pinheiro2015learning, de2017semantic} learned to segment instances by proposing segmentation candidates based on the region-based CNN (R-CNN) \cite{girshick2014rich}.
A top-down detector-based Mask R-CNN framework was first introduced by He et al. \cite{he2017mask} to simultaneously perform mask and class label prediction.
By contrast, bottom-up methods such as \cite{newell2017associative, liu2017sgn} aim to assign per-pixel predictions to instances. \\

\begin{figure*}[t]
 \begin{center}
  \includegraphics[width=17cm]{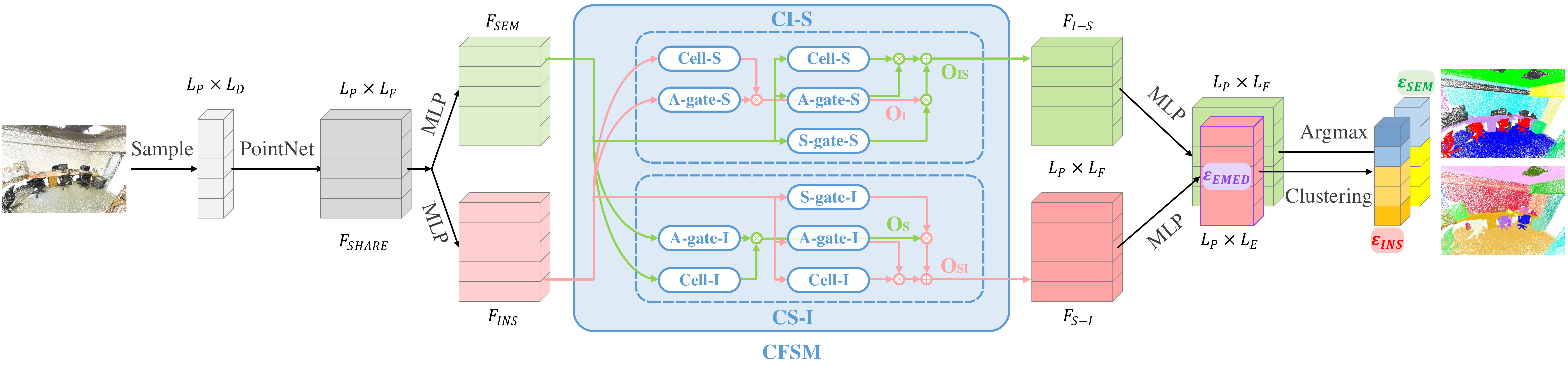}
 \end{center}
 \caption{Illustration of our proposed 3DCFS architecture. The randomly sampled point clouds are first input to a feed-forward network, which computes a 128-dimensional feature vector for each point. Then, the CFSM exploits and integrates the reciprocal information for semantic and instance segmentations. $\mathcal{E}_{EMED}$ is applied on the instance embeddings to help the model learn to balance the magnitudes of the embedding dimensions during training. }
 \label{fig2}
\end{figure*}

\noindent \textbf{3D Point Clouds Segmentation.}
Recent advances in deep neural networks have also led to various cutting-edge 3D semantic \cite{guerry2017snapnet, shi2015deeppano, su2015multi, qi2016volumetric, komarichev2019cnn, wu2018pointconv, ye20183d} and instance segmentation \cite{lahoud20193d, hou20193d, yi2019gspn, wang2018sgpn} approaches.
Using voxelized volumes to represent 3D point clouds is a popular and effective strategy. \cite{maturana2015voxnet, wu20153d, huang2016point, riegler2017octnet} transferred 3D point clouds data into regular volumetric occupancy grids and applied 3D CNNs to perform voxel-level predictions.
Based on the MLP, PointNet \cite{qi2017pointnet} was the first to directly process raw point clouds and perform point-level predictions, demonstrating high performance on both segmentation and classification tasks.
Following that pioneering work, PointNet++ \cite{qi2017pointnet++}, PointCNN \cite{li2018pointcnn}, GB-RCU \cite{engelmann2017exploring} and RSNet \cite{huang2018recurrent} were developed through investigations of the local context and hierarchical learning structures.
Graph neural networks have opened up more efficient and flexible ways to handle 3D segmentation \cite{wang2018dynamic, landrieu2018large, wang2019graph}.
Recently, by advancing a joint semantic and instance learning framework, \cite{wang2019associatively, pham2019jsis3d} proposed methods that achieve superior performance on both tasks. \\

\section{Method}
\label{sec:method}
As depicted in Figure \ref{fig2}, the framework with CFSM and $\mathcal{E}_{EMED}$ removed is the baseline method.
First, point clouds of size $L_P$ are encoded into a feature matrix $F_{SHARE} \in \mathbb{R}^{L_P \times L_F}$ by the encoder (PointNet/PointNet++).
Next, two tasks separately decode the shared encoded feature for their own missions. $F_{SHARE}$ is decoded by the semantic segmentation branch into the semantic feature matrix $F_{SEM} \in \mathbb{R}^{L_P \times L_F}$ and then outputs the semantic predictions $P_{SEM} \in \mathbb{R}^{L_P \times L_C}$, where $L_C$ is the semantic class number.
The instance segmentation branch decodes $F_{SHARE}$ into the instance feature matrix $F_{INS} \in \mathbb{R}^{L_P \times L_F}$, which is utilized to predict the per-point instance embeddings $E_{INS} \in \mathbb{R}^{L_P \times L_E}$, where $L_E$ denotes the length of the output embedding dimensions. These embeddings are used to calculate the Euclidean distances between points for instance clustering.
During the training process, the semantic branch is supervised by cross-entropy loss and the instance branch is supervised by
the instance loss following \cite{wang2019associatively}, and the specific loss formula is detailed in \cite{wang2019associatively}. In our paper, we denote this loss as $\mathcal{E}_{INS}$.
For inference, we use mean-shift clustering \cite{comaniciu2002mean} on the instance embeddings to obtain the final instance labels.
The mode of the semantic labels of the points within the same instance is assigned as the predicted semantic class.

\subsection{CFSM}
\noindent \textbf{Reciprocal Feature Selection and Integration.}
As illustrated in Figure \ref{fig2}, our CFSM contains two branches: $\mathcal{C}_{I-S}$ for instance-fused semantic segmentation and $\mathcal{C}_{S-I}$ for semantic-aware instance segmentation.
Both $\mathcal{C}_{I-S}$ and $\mathcal{C}_{S-I}$ can be separately integrated into baseline model, when the other branch is replaced by the MLP.
In our method, there are two types of gates: attention gates (A-gates) and selection gates (S-gates). Both are learnable modules that implement several 1 $\times$ 1 convolutions and activation functions. An A-gate is used for reweighting the semantic and instance features themselves before fusion, while the S-gate is used to filter or select the information from the other task. The cell is a decoding unit that contains convolutions and activation functions. \\

\noindent \textbf{CI-S.} As depicted in Figure \ref{fig2}, we denote $F_{INS}$ and $F_{SEM}$ as the instance and semantic features decoded by the MLP from $F_{SHARE}$. The semantic branch $\mathcal{C}_{I-S}$ contains three units: an A-gate called ``A-gate-S'', an S-gate named ``S-gate-S'' and a cell termed ``Cell-S''. The units with the same name share the weight parameters.
For the $\mathcal{C}_{I-S}$ branch, the red $F_{INS}$ pass through the Cell-S and A-gate-S to obtain the output $O_I$; the $F_{SEM}$ in green  are fed into all three units to obtain the output $O_{IS}$. These outputs are calculated by the dot product $\otimes$ and summation $\oplus$ operations as illustrated in Figure \ref{fig2}, which are formulated as follows: \\
\begin{numcases}{}
 O_{I} = \sigma(W_{cell} * F_{INS}) \cdot \varsigma(W_{A} * F_{INS})  \\
 O_{i} = \xi(W_{S} * F_{SEM} \cdot O_{I}) \\
 O_{IS} = \sigma(W_{cell} * F_{SEM}) \cdot \varsigma(W_{A} * F_{SEM}) + O_{i} \\
 F_{I-S} \triangleq O_{IS}
\end{numcases}

where $W_{cell}$, $W_{A}$ and $W_{S}$ are the weights of Cell-S, A-gate-S and S-gate-S, respectively, and $\sigma$, $\varsigma$ and $\xi$ indicate their activation functions. The symbol $*$ denotes the convolution operation, and $\cdot$ represents the dot product.
The final output of $\mathcal{C}_{I-S}$ for the semantic segmentation task is $F_{I-S} \in \mathbb{R}^{L_P \times L_F}$.
Based on the attention mechanism, our A-gate has the capability to reweight the features of the task itself to facilitate extracting the crucial internal information of both tasks. Then, the representations in $F_{INS}$, which are useful to the semantic segmentation task, are exploited and reserved through our S-gate. This selection process is guided and controlled by the semantic features.
For example, the S-gate-S in $\mathcal{C}_{I-S}$ can select features from the same instances and discover their general characteristics, which helps the model to recognize their category. \\

\noindent \textbf{CS-I.} The $\mathcal{C}_{S-I}$ architecture is the same as the $\mathcal{C}_{I-S}$ structure, except that the instance features are assisted by the semantic features. $F_{SEM}$ is passed to $\mathcal{C}_{S-I}$ as complementary information to help improve the performance of the instance segmentation.
S-gate-I in $\mathcal{C}_{S-I}$ is able to block the useless information and filter the features that blur the differences between instances as well as select more valuable representations to indicate the differences between categories for instance segmentation.
The output of $\mathcal{C}_{S-I}$ for the semantic segmentation task is $F_{S-I} \in \mathbb{R}^{L_P \times L_F}$.
\\

\noindent \textbf{CFSM.} As illustrated in Figure \ref{fig2}, $\mathcal{C}_{I-S}$ and $\mathcal{C}_{S-I}$ can be simultaneously integrated into the baseline model.
As the complementary information, $F_{SEM}$ and $F_{INS}$ are taken as the input to the other branches $\mathcal{C}_{S-I}$ and $\mathcal{C}_{I-S}$, respectively. At the same time, $F_{SEM}$ and $F_{INS}$ are passed to their own branches $\mathcal{C}_{I-S}$ and $\mathcal{C}_{S-I}$, respectively. \\

\subsection{Learn to Balance the Embedding Dimension Magnitude}
\label{trick}
To further improve the instance segmentation performance of our framework, we design a loss function to learn to balance the magnitudes of the output embedding dimensions. The Euclidean distance is sensitive to magnitude differences, which makes the cluster results dependent on only a few dimensions of the embedding and reduces the generalizability of the model.
A traditional trick to solve this issue is to apply a mean-removal strategy on the output embeddings before instance clustering during inference.
Rather than employ this post-process method, we directly apply our proposed loss function to the model to stabilize the Euclidean distance calculation during training. Specifically, we denote $\mathcal{E}_{EMED}$ as the equilibrium loss for the magnitude. The loss term can be written as follows:
\begin{numcases}{}
 \bar{E} = \frac{1}{L_P} \sum_{i=1}^{L_P}E_i                    \\
 \mathcal{E}_{EMED} = \frac{1}{L_E} \sum_{d=1}^{L_E} (\bar{E}_{d}-\upmu)^2 \\
 \mathcal{E}_{INS}^{*} = \mathcal{E}_{INS}+ \alpha \mathcal{E}_{EMED}
\end{numcases}
where $E_i$ is the embedding of each point, $\mathcal{E}_{INS}^{*}$ is the total instance loss of our 3DCFS, $\upmu$ denotes the mean value of $\bar{E}$, and $\alpha$ is the balanced weight of $\mathcal{E}_{INS}$ and $\mathcal{E}_{EMED}$.

\section{Experiments}
\label{sec:eval}
\subsection{Datasets and Experimental Setup}
\noindent \textbf{Dataset.} Our experiments are conducted on two benchmark datasets: the Stanford 3D Indoor Semantics Dataset (S3DIS) \cite{armeni20163d} and ShapeNet Dataset\cite{yi2016scalable}.
\begin{itemize}
 \item S3DIS is a 3D scene dataset that contains large-scale scans of indoor spaces. Each point is annotated with an instance label and a semantic label from 13 semantic classes. S3DIS embeds each point into a 9-dimensional feature vector including XYZ, RGB and normalized coordinates. Following \cite{qi2017pointnet}, we split the rooms into 1 m $\times$ 1 m overlapping blocks with stride 0.5 m on the ground plane and sample 4,096 points from each block.
 \item The ShapeNet part dataset contains 16,881 3D shapes from 16 semantic classes. Each point is associated with one of the 50 different parts. We utilize the instance annotations from \cite{wang2018sgpn} as the ground-truth labels. Each shape is represented by point clouds with 2,048 points following \cite{qi2017pointnet}, and each point is represented by an XYZ 3-dimensional vector. The point clouds are sampled for the input of our framework following \cite{wang2019associatively}.
\end{itemize}

\noindent \textbf{Evaluation.} Following \cite{wang2019associatively}, we conduct experiments involving S3DIS on Area5.
The performance on the sixth fold cross validation with microaveraging \cite{de2017semantic} is also measured. For semantic segmentation, we calculate the overall accuracy (oAcc), mean accuracy (mAcc) and mean IoU (mIoU).
To evaluate the performance of instance segmentation, we use the coverage (Cov) and weighted coverage (WCov), the specific calculation formulas are detailed in \cite{ren2017end, liu2017sgn, zhuo2017indoor}. \\

\noindent \textbf{Implementation Details.}
For instance segmentation, we train 3DCFS with $\lambda = 0.001$. We use five output embeddings following \cite{wang2019associatively} and set $\alpha$ to 0.01.
We train the network for 50 epochs for PointNet and PointNet++ with a batch size of 12 and the base learning rate set to 0.001 and divided by 2 every 300 k iterations.
We select the Adam optimizer to optimize the network on a single GPU (Tesla P100) and set the momentum to 0.9 for the training process.
During the inference process, we set the bandwidth to 0.6 for mean-shift clustering and apply the BlockMerging algorithm \cite{wang2018sgpn} to merge instances from different blocks. The code will be available at GitHub, which contains more details.

\subsection{S3DIS}
Following \cite{wang2019associatively}, we conducted experiments on the S3DIS dataset based on the PointNet and PointNet++ backbone networks. \\

\begin{table}[htb]
 \caption{Instance (red) and semantic (green) segmentation results on S3DIS dataset (Test on Area5).}
 \footnotesize
 \centering
 \resizebox{84mm}{20mm}{
  \begin{tabular}{c|l|c|c|c|c}
   \toprule
   \hline
   \rowcolor{mypink}
   Backbone \cellcolor{mygray} & Method \cellcolor{mygray}         & mCov                     & mWCov                      & mPrec                     & mRec             \\
   \hline
   \multirow{3}{*}{PN}         & SGPN \cite{wang2018sgpn}          & 32.7                     & 35.5                       & 36.0                      & 28.7             \\
                               & ASIS \cite{wang2019associatively} & 38.2                     & 41.6                       & 44.2                      & 35.6             \\
                               & 3DCFS                             & \textbf{41.2}            & \textbf{44.4}              & \textbf{47.5}             & \textbf{39.4}    \\
   \hline
   \multirow{2}{*}{PN++}
                               & ASIS \cite{wang2019associatively} & 44.7                     & 47.6                       & 54.3                      & 43.2             \\
                               & 3DCFS                             & \textbf{49.0}            & \textbf{52.1}              & \textbf{55.5}             & \textbf{45.9}    \\
   \hline
   Backbone \cellcolor{mygray} & Method \cellcolor{mygray}         & mAcc \cellcolor{mygreen} & mIoU   \cellcolor{mygreen} & oAcc  \cellcolor{mygreen} & \multirow{6}*{/} \\
   \cline{0-4}
   \multirow{3}{*}{PN}         & PN \cite{qi2017pointnet}          & 52.1                     & 43.4                       & 83.5                                         \\
                               & ASIS \cite{wang2019associatively} & 55.4                     & 46.5                       & 84.8                                         \\
                               & 3DCFS                             & \textbf{56.4}            & \textbf{47.1}              & \textbf{84.9}                                \\
   \cline{0-4}
   \multirow{2}{*}{PN++}
                               & ASIS \cite{wang2019associatively} & 60.9                     & 53.4                       & 86.9                                         \\
                               & 3DCFS                             & \textbf{62.7}            & \textbf{55.5}              & \textbf{87.8}                                \\
   \bottomrule
  \end{tabular}}
 \label{tab1}
\end{table}

\begin{table}[htb]
 \caption{Instance (red) and semantic (green) segmentation results on S3DIS dataset (Test on 6-fold CV).}
 \footnotesize
 \centering
 \resizebox{84mm}{20mm}{
  \begin{tabular}{c|l|c|c|c|c}
   \toprule
   \hline
   \rowcolor{mypink}
   Backbone \cellcolor{mygray} & Method \cellcolor{mygray}         & mCov                     & mWCov                      & mPrec                     & mRec             \\
   \hline
   \multirow{3}{*}{PN}         & SGPN \cite{wang2018sgpn}          & 37.9                     & 40.8                       & 38.2                      & 31.2             \\
                               & ASIS \cite{wang2019associatively} & 44.7                     & 48.4                       & 53.7                      & 41.0             \\
                               & 3DCFS                             & \textbf{46.1}            & \textbf{49.8}              & \textbf{55.5}             & \textbf{42.7}    \\
   \hline
   \multirow{2}{*}{PN++}
                               & ASIS \cite{wang2019associatively} & 51.5                     & 54.8                       & 62.8                      & 47.2             \\
                               & 3DCFS                             & \textbf{53.1}            & \textbf{57.1}              & \textbf{63.7}             & \textbf{49.1}    \\
   \hline
   Backbone \cellcolor{mygray} & Method \cellcolor{mygray}         & mAcc \cellcolor{mygreen} & mIoU   \cellcolor{mygreen} & oAcc  \cellcolor{mygreen} & \multirow{6}*{/} \\
   \cline{0-4}
   \multirow{3}{*}{PN}         & PN \cite{qi2017pointnet}          & 60.3                     & 48.9                       & 80.3                                         \\
                               & ASIS \cite{wang2019associatively} & 62.9                     & 51.6                       & 82.0                                         \\
                               & 3DCFS                             & \textbf{63.8}            & \textbf{52.3}              & \textbf{82.5}                                \\
   \cline{0-4}
   \multirow{2}{*}{PN++}
                               & ASIS \cite{wang2019associatively} & 70.1                     & 59.3                       & 86.2                                         \\
                               & 3DCFS                             & \textbf{72.4}            & \textbf{60.3}              & \textbf{86.3}                                \\
   \bottomrule
  \end{tabular}}
 \label{tab2}
\end{table}

\begin{figure}[t]
 \begin{center}
  \includegraphics[width=8.4cm]{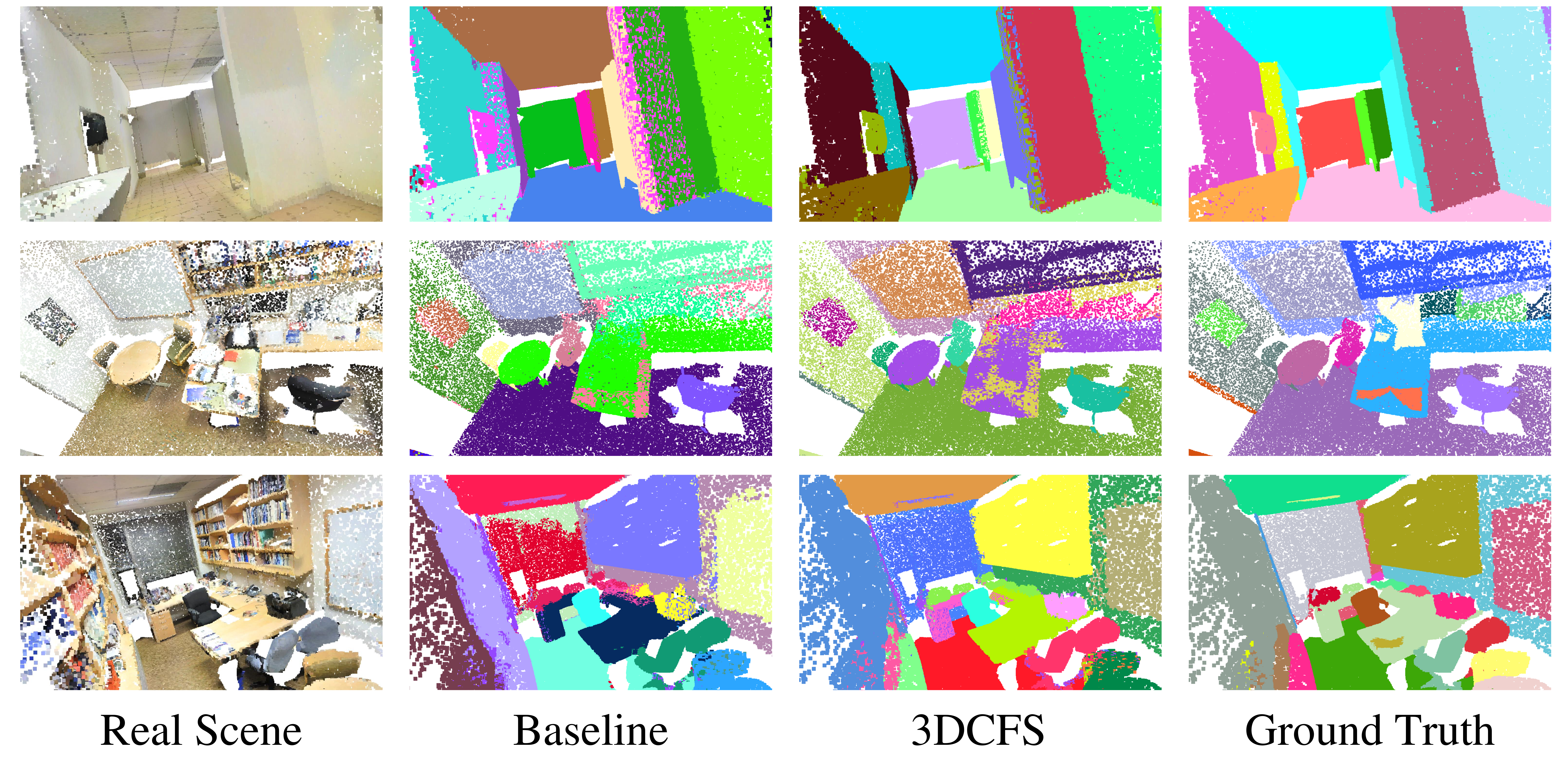}
 \end{center}
 \caption{Comparison of the baseline method and 3DCFS on instance segmentation. The different colors represent different instances.}
 \label{fig3}
\end{figure}
\begin{figure}[t]
 \begin{center}
  \includegraphics[width=8.4cm]{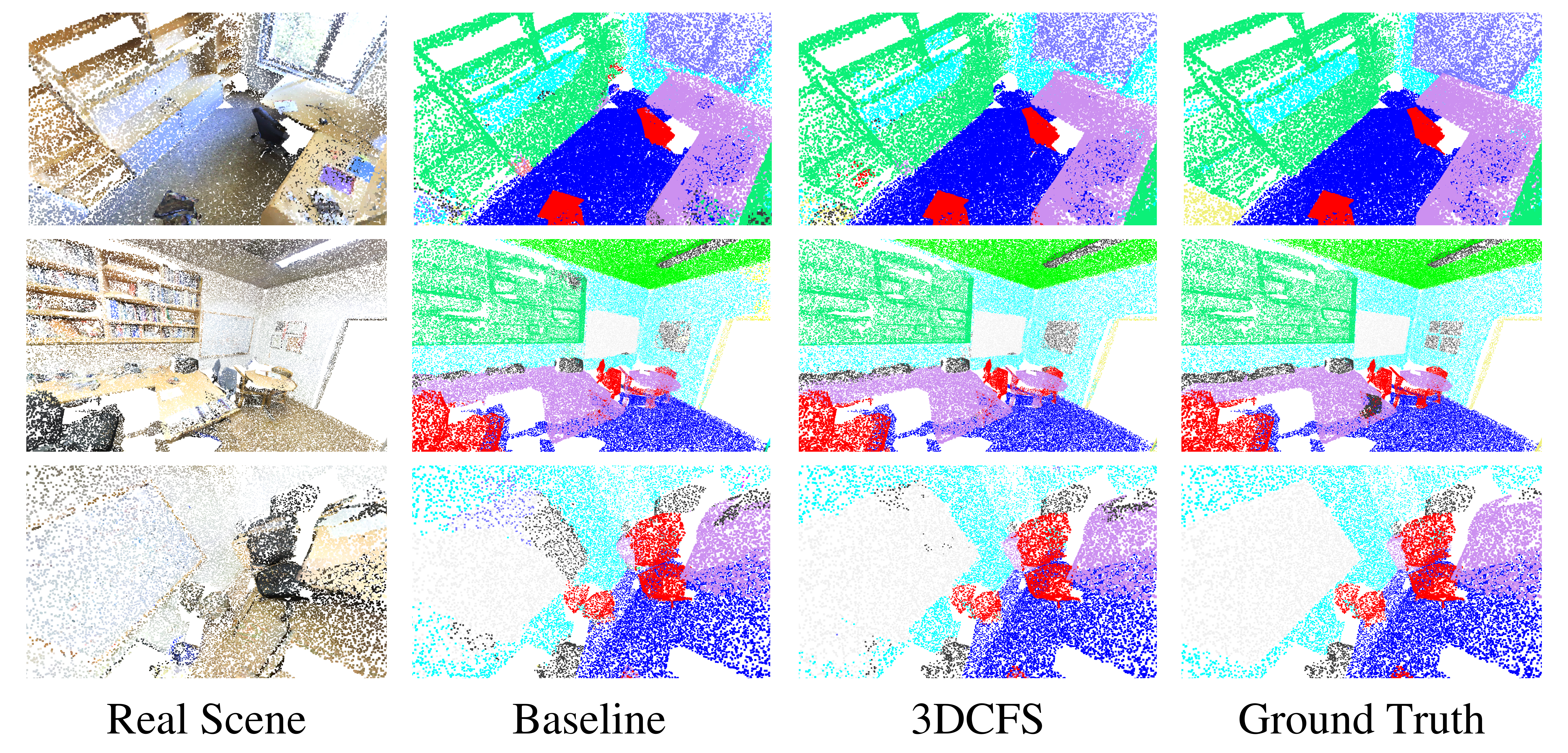}
 \end{center}
 \caption{Comparison of the baseline method and 3DCFS on semantic segmentation.}
 \label{fig4}
\end{figure}
\begin{figure}
 \begin{center}
  \centerline{\includegraphics[width=9.2cm]{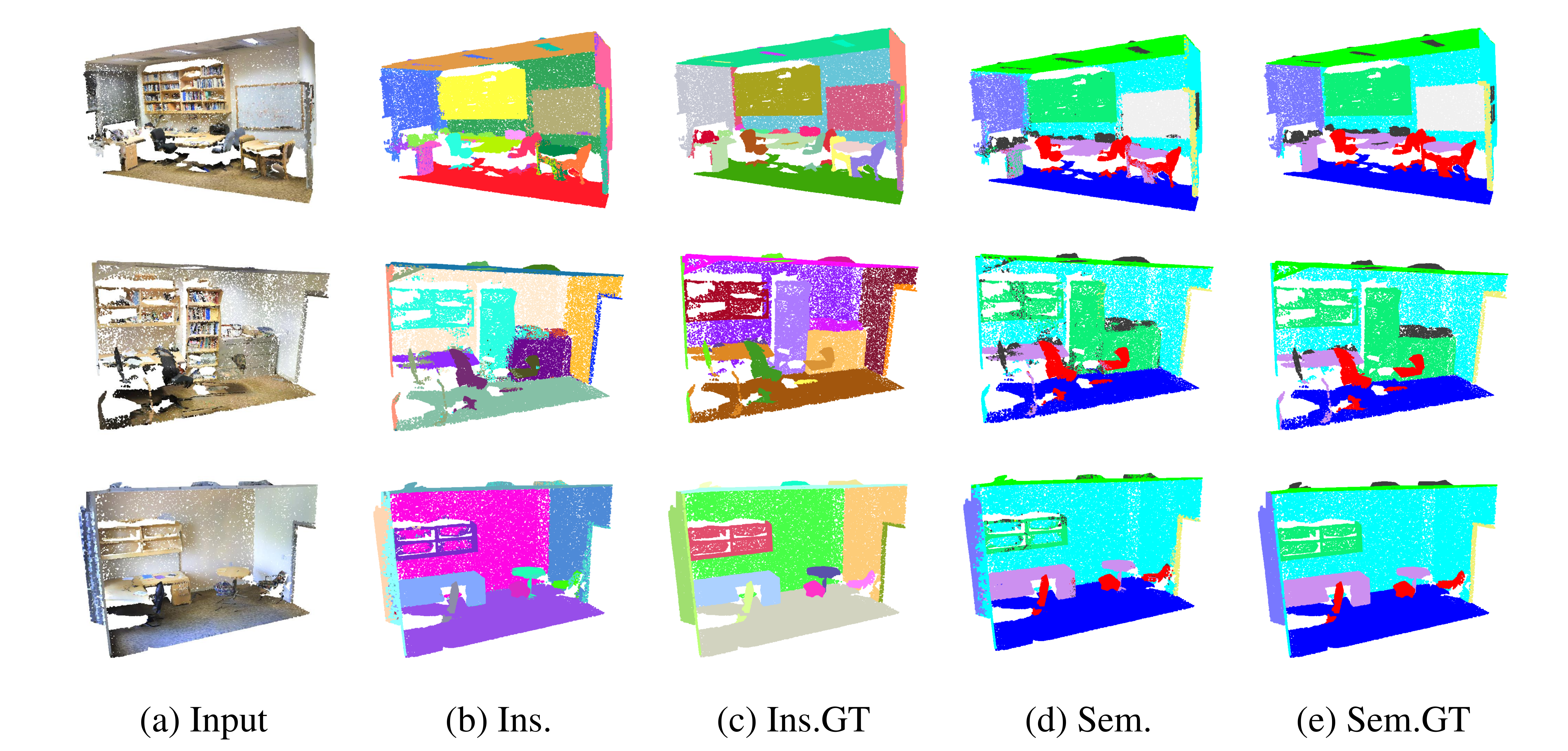}}
 \end{center}
 \caption{Qualitative results of 3DCFS on the S3DIS test fold.}
 \label{fig5}
\end{figure}

\noindent \textbf{Quantitative Results.}
For Area5, the quantitative results of 3DCFS on the instance and semantic segmentation tasks are shown in Table \ref{tab1}. Using the PointNet backbone, 3DCFS achieves 44.4 mWCov, which dramatically outperforms the state-of-the-art method ASIS\cite{wang2019associatively} by 2.8 and significantly improves the mPrec by 3.3.
After replacing the backbone with PointNet++, we still achieve 4.5 mWCov gains and 2.1 mIoU gains on the instance and semantic segmentation tasks, respectively.
Table \ref{tab2} shows the performance of our 3DCFS on semantic segmentation in all 6 areas.
3DCFS outperforms ASIS by 1.4 for mWCov and 1.8 for mPrec.
Using the PointNet++ backbone on all 6 areas, 3DCFS improves the mAcc by 2.3 and the mIou by 1.0.
Clearly, our 3DCFS method outperforms the SOTA method ASIS\cite{wang2019associatively} by a large margin.
As reported in Table \ref{tab1} and \ref{tab2}, whether constructed upon the PointNet or PointNet++ backbones, evaluated in Area5 or 6-fold CV, our method consistently obtains better performance on both instance and semantic segmentation tasks than the state-of-the-art methods.
The stable improvement demonstrates that our 3DCFS is a general and effective framework that can be built upon different network backbones.
Table \ref{tab4} shows the instance and semantic segmentation results for specific categories.
We reproduced the results of ASIS \cite{pham2019jsis3d} and JSIS3D \cite{wang2019associatively} using the code at GitHub published by the respective authors to make a full class comparison with the same PointNet backbone. \\

\begin{table}[htb]
 \caption{Ablation study on the S3DIS dataset in Area5. CI-S refers to only instance fusion; CS-I refers to only semantic awareness; CFSM contains both CI-S and CS-I; CFSM (post) means CFSM with post-process (mean-removal) on instance embedding; 3DCFS is our full approach equipped with $\mathcal{E}_{EMED}$.}
 \footnotesize
 \centering
 \resizebox{84mm}{12mm}{
  \begin{tabular}{p{5.5em}|p{2em}p{3.0em}p{2.5em}p{2em}|p{2em}p{2em}p{2em}}
   \toprule
   \hline
   Method \cellcolor{mygray} & mCov \cellcolor{mypink} & mWCov \cellcolor{mypink} & mPrec \cellcolor{mypink} & mRec \cellcolor{mypink} & mAcc \cellcolor{mygreen} & mIou \cellcolor{mygreen} & oAcc \cellcolor{mygreen} \\
   \hline
   Baseline                  & 46.0                    & 49.0                     & 51.3                     & 42.0                    & 61.0                     & 52.9                     & 86.6                     \\
   CI-S                      & 46.7                    & 49.6                     & 53.9                     & 43.1                    & 62.2                     & 54.4                     & 87.4                     \\
   CS-I                      & 47.0                    & 50.1                     & 54.5                     & 43.3                    & 61.6                     & 53.9                     & 87.2                     \\
   CFSM                      & 48.0                    & 50.8                     & 54.7                     & 44.6                    & 62.3                     & 54.5                     & 87.7                     \\
   CFSM (post)               & 48.4                    & 51.6                     & \textbf{55.6}            & 44.0                    & 62.4                     & 54.6                     & 87.5                     \\
   3DCFS                     & \textbf{49.0}           & \textbf{52.1}            & 55.5                     & \textbf{45.9}           & \textbf{62.7}            & \textbf{55.5}            & \textbf{87.8}            \\
   \bottomrule
  \end{tabular}}
 \label{tab3}
\end{table}
\begin{table*}[htb]
 \caption{Per class results on the S3DIS dataset.}
 \footnotesize
 \centering
 \resizebox{170mm}{13mm}{
  \begin{tabular}{c|c|c|ccccccccccccc}
   \toprule
   \hline
   Metrics \cellcolor{mygray} & Method                              & mean          & ceiling       & floor         & wall          & beam          & column        & window        & door          & table         & chair         & sofa          & bookcase      & board         & clutter       \\
   \hline
   \cellcolor{mypink}         & JSIS3D \cite{pham2019jsis3d}        & 41.5          & 82.7          & 85.1          & 44.2          & 0.0           & 15.4          & 74.4          & 33.3          & 34.0          & \textbf{64.3} & \textbf{20.0} & \textbf{47.5} & 9.1           & 30.3          \\
   \cellcolor{mypink}  mPrec  & ASIS \cite{wang2019associatively}   & 53.7          & \textbf{88.6} & 87.5          & 57.8          & 57.1          & 35.7          & \textbf{68.3} & 59.1          & \textbf{43.2} & 58.4          & 7.3           & 36.0          & 64.8          & 34.4          \\
   \cellcolor{mypink}         & 3DCFS                               & \textbf{55.3} & 88.1          & \textbf{89.1} & \textbf{59.3} & \textbf{63.3} & \textbf{41.1} & \textbf{68.3} & \textbf{61.2} & 42.2          & 55.5          & 4.3           & 38.4          & \textbf{73.4} & \textbf{34.6} \\
   \hline
   \cellcolor{mygreen}        & SEGCloud \cite{tchapmi2017segcloud} & 48.9          & 90.1          & 96.1          & 69.9          & 0.0           & 18.4          & 38.4          & 23.1          & 70.4          & \textbf{75.9} & \textbf{40.9} & 58.4          & 13.0          & 41.6          \\
   mAcc \cellcolor{mygreen}   & JSIS3D \cite{pham2019jsis3d}        & 50.5          & \textbf{96.7} & \textbf{99.1} & \textbf{90.0} & 0.0           & 0.1           & 53.0          & 10.2          & 64.3          & 71.2          & 35.3          & \textbf{64.6} & 17.3          & 54.5          \\
   \cellcolor{mygreen}        & ASIS \cite{wang2019associatively}   & 62.9          & 95.6          & 91.5          & 88.4          & 62.4          & 37.1          & \textbf{57.8} & 67.0          & \textbf{77.5} & 55.7          & 23.2          & 53.1          & 43.0          & 65.2          \\
   \cellcolor{mygreen}        & 3DCFS                               & \textbf{64.7} & 94.9          & 92.1          & 88.4          & \textbf{64.7} & \textbf{44.8} & 56.7          & \textbf{69.2} & 74.2          & 56.4          & 34.3          & 53.1          & \textbf{45.8} & \textbf{66.8} \\
   \bottomrule
  \end{tabular}}
 \label{tab4}
\end{table*}

\begin{table}[htb]
 \caption{Ablation study on the S3DIS dataset in Area5. CFSM5 and CFSM10 represent the instance segmentation with output embedding length of 5 and 10, respectively. 3DCFS5 and 3DCFS10 denote the method equipped with $\mathcal{E}_{EMED}$.}
 \footnotesize
 \centering
 \resizebox{84mm}{9mm}{
  \begin{tabular}{p{4em}|p{2em}p{3.0em}p{2.5em}p{2em}|p{2em}p{2em}p{2em}}
   \toprule
   \hline
   Method \cellcolor{mygray} & mCov \cellcolor{mypink} & mWCov \cellcolor{mypink} & mPrec \cellcolor{mypink} & mRec \cellcolor{mypink} & mAcc \cellcolor{mygreen} & mIou \cellcolor{mygreen} & oAcc \cellcolor{mygreen} \\
   \hline
   CFSM5                     & 48.0                    & 50.8                     & 54.7                     & 44.6                    & 62.3                     & 54.5                     & 87.7                     \\
   3DCFS5                    & 49.0                    & 52.1                     & 55.5                     & 45.9                    & 62.7                     & 55.5                     & 87.8                     \\
   \hline
   CFSM10                    & 45.4                    & 48.5                     & 52.2                     & 40.5                    & 61.2                     & 53.6                     & 87.3                     \\
   3DCFS10                   & 48.4                    & 51.7                     & 55.0                     & 44.1                    & 63.1                     & 55.2                     & 87.5                     \\
   \bottomrule
  \end{tabular}}
 \label{tab8}
\end{table}

\noindent \textbf{Ablation Study.}
The ablation study results are shown in Table \ref{tab3}. Compared with the baseline, our 3DCFS achieves obvious improvements. We obtain 3.1 mWCov and 4.2 mPrec gains for the instance segmentation task. For semantic segmentation, we achieve 1.7 mAcc and 2.6 mIou gains.
Specifically, as shown in Table \ref{tab3}, equipped with only CI-S, our method achieves 54.4 mIoU and 62.2 mAcc, which outperforms the baseline by 1.5 and 1.2 on the semantic segmentation task, respectively.
Adopting only CS-I yields 50.1 mWCov and 54.5 mPrec, which contributes to a 1.1 gain in mWCov and a 3.2 gain in mPrec compared to the baseline on the instance segmentation task.
Comparing CS-I and CI-S, we find that CS-I outperforms CI-S on instance metrics, while CI-S performs better on the semantic task.
The coupled module CFSM further outperforms the baseline results by a large performance margin, achieving 62.3 mAcc (semantic) and 54.7 mPrec (instance), both of which are larger than the gains provided by the individual CS-I and CI-S.
The results demonstrate that improving one task can also help improve the other because each task learns better reciprocal features.

\begin{figure}
 \begin{center}
  \centerline{\includegraphics[width=6.2cm]{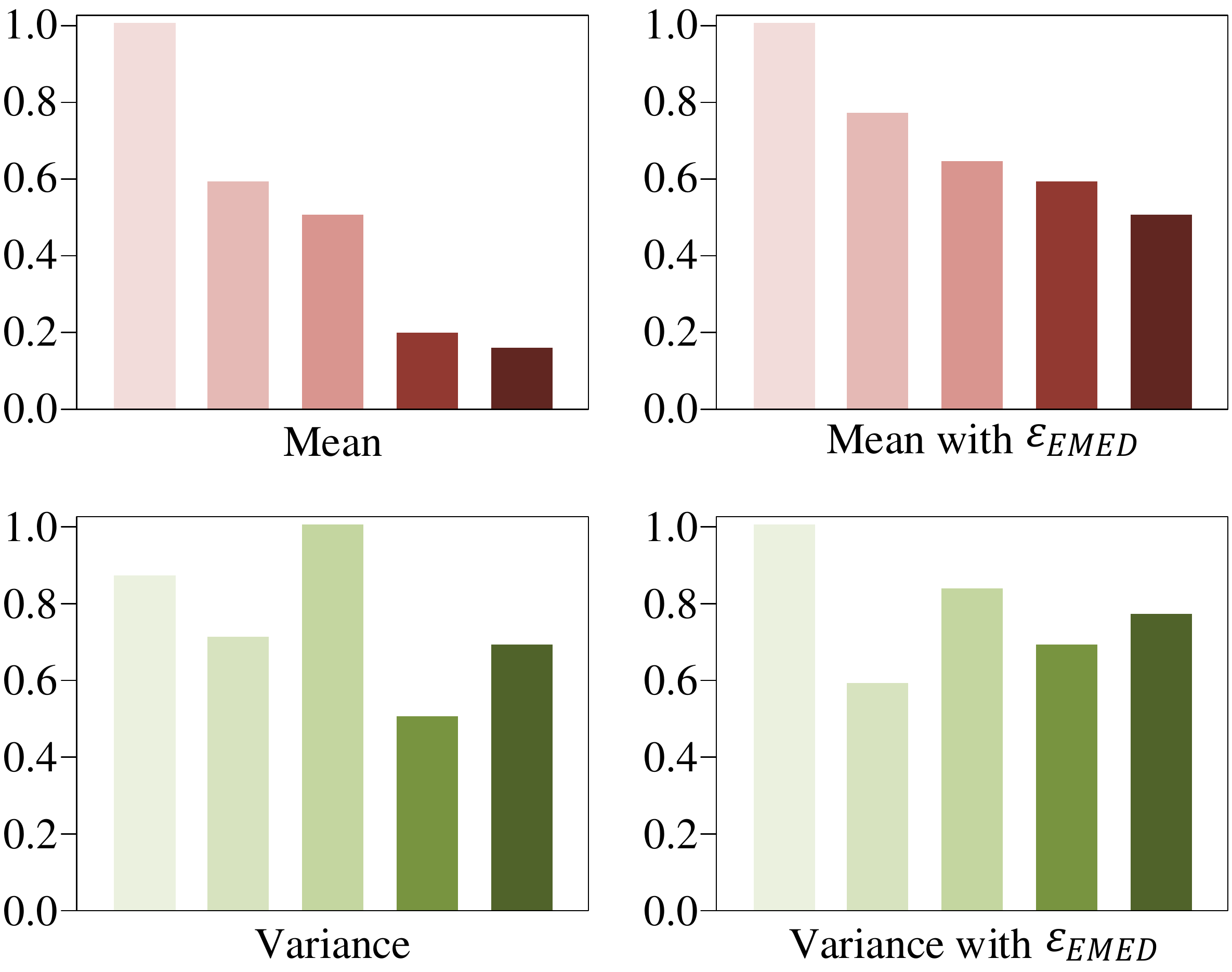}}
 \end{center}
 \caption{Comparison of the mean and variance of 5 embedding dimensions with or without $\mathcal{E}_{EMED}$. All the values are normalized, and the mean values are ranked from high to low for better visualization.}
 \label{fig6}
\end{figure}

Table \ref{tab3} also compares our $\mathcal{E}_{EMED}$ with the post-processing method mentioned in Section \ref{trick}.
Figure \ref{fig6} shows the comparison of the mean and variance of 5 dimensions of the embeddings with or without $\mathcal{E}_{EMED}$. The statistical analysis of the results reveals that the mean values are balanced and the variances are not influenced by $\mathcal{E}_{EMED}$, which indicates that it maintains the representational ability of the instance feature.
The superior performance shows that our $\mathcal{E}_{EMED}$ successfully helps the model learn to balance the dimension magnitude.
Table \ref{tab8} shows that incorporating our proposed $\mathcal{E}_{EMED}$ boosts the performance with embedding lengths 5 and 10. Note that the improvement is much more significant as the embedding length increases. \\

\noindent \textbf{Qualitative Results.}
For instance segmentation, different colors represent different instances. As depicted in Figure \ref{fig3}, the baseline approach incorrectly clusters two nearby different class instances into one instance (e.g., board and wall).
After applying 3DCFS, the instances are correctly clustered.
For semantic segmentation, each color refers to a particular class. The qualitative comparisons are shown in Figure \ref{fig4}.
3DCFS performs better on classifying the entire semantic information, especially at the boundaries of different categories.
Figure \ref{fig5} shows qualitative examples of 3DCFS on both instance and semantic segmentation.
Our results are essentially the same as the ground truth, especially for instance segmentation. \\

\noindent \textbf{Speed and Computing Resources.}
Table \ref{tab5} shows a comparison of the memory cost and computation time measured on a single GTX 1080 GPU. For a fair comparison, we conducted the experiments in the same environment, including the same GPU, batch size (4) and data (Area5).
Note that all time units are minutes and all memory units are MB. In the training process, the result is the time and memory cost for one epoch. Our approach takes only 26.4 minutes and 2,227 MB, which is significantly faster and more efficient than the state-of-the-art methods.
In the test process, the results show the resource consumption for inferencing. Here, our method is also found to be superior to the state-of-the-arts in terms of accuracy, speed and computational cost.
\begin{table}[h]
 \caption{Comparisons of computation speed, GPU memory and performance. The units for time and memory are minutes and MB respectively.}
 \footnotesize
 \centering
 \resizebox{84mm}{10mm}{
  \begin{tabular}{c|cc|cc|c}
   \toprule
   \multirow{2}{*}{\diagbox{Method}{Metrics}} & \multicolumn{2}{c|}{Train} & \multicolumn{2}{c|}{Test} & \multirow{2}{*}{mWconv}                                \\
   \cline{2-5}
                                              & time                       & memory                    & time                    & memory                       \\
   \hline
   SGPN \cite{wang2018sgpn}                   & 59.3                       & 7549                      & 209.5                   & 420          & 35.5          \\
   ASIS \cite{wang2019associatively}          & 64.7                       & 4275                      & 54.2                    & 1235         & 40.3          \\
   3DCFS                                      & \textbf{26.4}              & \textbf{2227}             & \textbf{36.3}           & \textbf{307} & \textbf{44.4} \\
   \bottomrule
  \end{tabular}}
 \label{tab5}
\end{table}

\begin{table}[htb]
 \caption{Semantic segmentation results on ShapeNet datasets.}
 \footnotesize
 \centering
 \begin{tabular}{c|c}
  \toprule
  \hline
  Method   \cellcolor{mygray}       & mIoU \cellcolor{mygreen} \\
  \hline
  PointNet \cite{qi2017pointnet++}  & 84.3                     \\
  \hline
  ASIS \cite{wang2019associatively} & 85.0                     \\
  SGPN \cite{wang2018sgpn}          & 85.8                     \\
  \hline
  3DCFS                             & \textbf{87.6}            \\
  \bottomrule
 \end{tabular}
 \label{tab6}
\end{table}

\begin{figure}
 \begin{center}
  \centerline{\includegraphics[width=8.5cm]{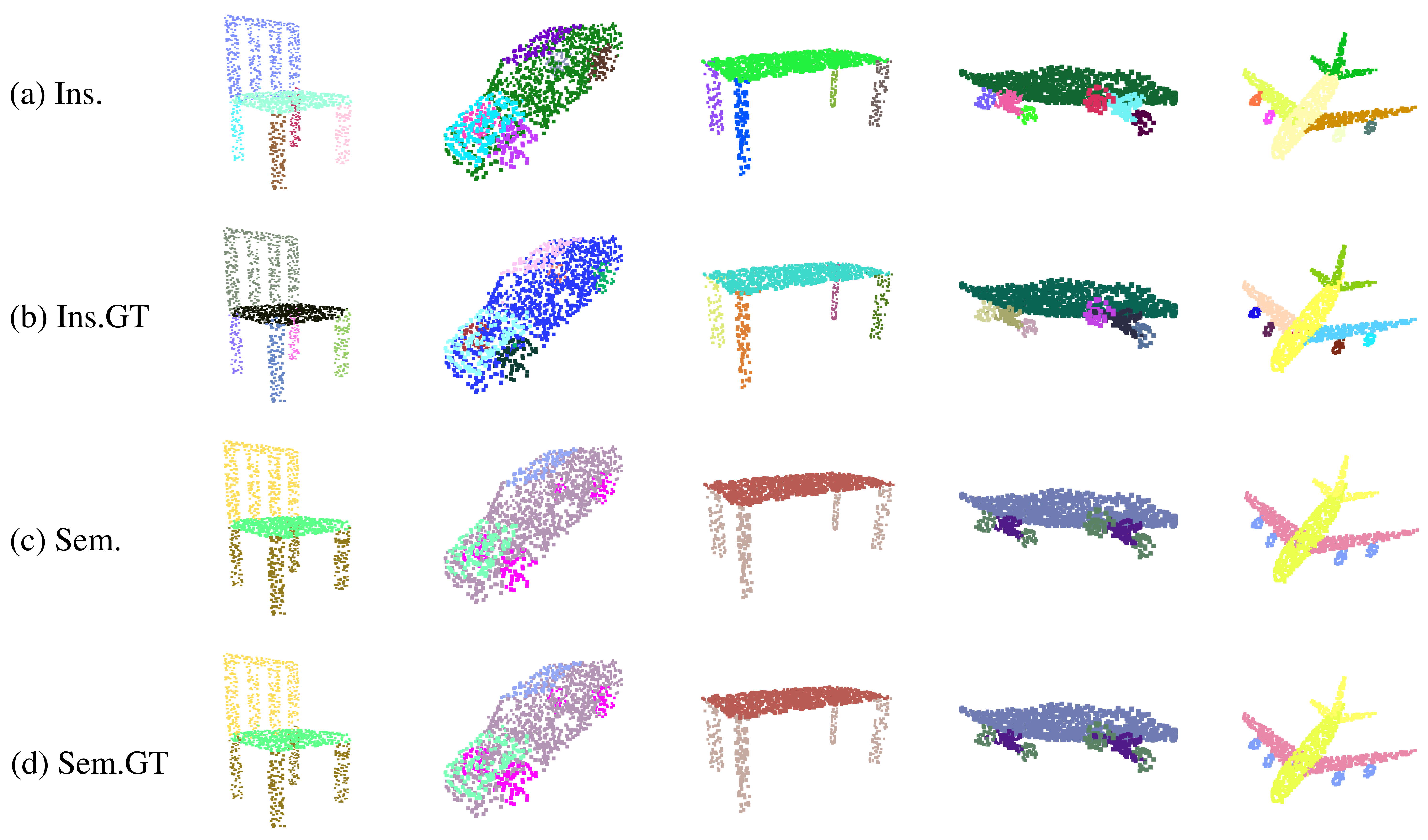}}
 \end{center}
 \caption{Qualitative results of 3DCFS on the ShapeNet test split. (a) Instance segmentation results of 3DCFS. (b) Generated ground truth for instance segmentation. (c) Semantic segmentation results of 3DCFS. (d) Semantic segmentation ground truth.}
 \label{fig7}
\end{figure}
\subsection{ShapeNet}
We conducted experiments on the ShapeNet dataset using instance segmentation annotations generated by \cite{wang2018sgpn}.
For instance segmentation, only the qualitative results are provided following \cite{wang2018sgpn} because no true ground truth exists.
As shown in Figure \ref{fig7}, the tires of the car and legs of the chair and the table are properly grouped into individual instances. Both the semantic and instance segmentation results are accurate and clear.
The semantic segmentation results are shown in Table \ref{tab6}.
Our 3DCFS further outperforms the state-of-the-art method SGPN by 1.8 mIoU based on PointNet++.
These results reveal that our proposed 3DCFS also has the capability to boost part segmentation performance.

\section{Conclusions}
\label{sec:conclusion}
In this paper, we proposed a fast and robust joint 3D semantic-instance segmentation framework.
A novel CFSM was introduced to exploit the reciprocal information from two different tasks in a coupled manner.
We also proposed a novel loss function that helped our 3DCFS learn to balance the magnitudes of the instance embedding dimensions to make the Euclidean distance calculation more reliable.
Experimental results on the S3DIS and ShapeNet part datasets demonstrated the effectiveness and efficiency of 3DCFS.

\bibliographystyle{IEEEtran}
\bibliography{dleam}

\end{document}